# Learning the Ordering of Coordinate Compounds and Elaborate Expressions in Hmong, Lahu, and Chinese

**Chenxuan Cui**      **Katherine J. Zhang**      **David R. Mortensen**
Language Technologies Institute, Carnegie Mellon University
cxcui@cs.cmu.edu, kjzhang@cmu.edu, dmortens@cs.cmu.edu

## Abstract

Coordinate compounds (CCs) and elaborate expressions (EEs) are coordinate constructions common in languages of East and Southeast Asia. Mortensen (2006) claims that (1) the linear ordering of EEs and CCs in Hmong, Lahu, and Chinese can be predicted via phonological hierarchies and (2) these phonological hierarchies lack a clear phonetic rationale. These claims are significant because morphosyntax has often been seen as in a feed-forward relationship with phonology, and phonological generalizations have often been assumed to be phonetically "natural". We investigate whether the ordering of CCs and EEs can be learned empirically and whether computational models (classifiers and sequence labeling models) learn unnatural hierarchies similar to those posited by Mortensen (2006). We find that decision trees and SVMs learn to predict the order of CCs/EEs on the basis of phonology, with DTs learning hierarchies strikingly similar to those proposed by Mortensen. However, we also find that a neural sequence labeling model is able to learn the ordering of elaborate expressions in Hmong very effectively without using any phonological information. We argue that EE ordering can be learned through two independent routes: phonology and lexical distribution, presenting a more nuanced picture than previous work. [ISO 639-3: hmn, lhu, cmn] [1]

## 1 Introduction

In many languages of East and Southeast Asia, there are common constructions in which two words or phrases are coordinated without an overt marker like a conjunction (Hanna, 2013; Filbeck, 1996; Johns and Strecker, 1987; Wheatley, 1982; Matisoff, 1973; Pan and Cao, 1972; Watson, 1966; Banker, 1964). In coordinate compounds (CCs), two words are combined to form a compound word whose semantics are often a generalization of those of the two conjoined words. Elaborate expressions (EEs) are similar, except that they can consist of two phrases (rather than words) and include a repeated word. Take the following examples:

(1) Chinese coordinate compounds (CCs)
    父母 *fùmǔ*    father-mother  'parents'
    花木 *huāmù*   flower-tree    'vegetation'
    天地 *tiāndì*  heaven-earth   'universe'
    国家 *guójiā*  country-home   'nation'
    风水 *fēngshuǐ* wind-water    'geomancy'

(2) Lahu elaborate expressions (EEs)
    a. ɔ̂   cē    ɔ̂    phɔ̂
       four corner four side
       'at every corner'
    b. chɔ   phôʔ chɔ   dì
       people pile people lump
       'a throng of people'
    c. câ  cûʔ   dɔ̀  cûʔ
       eat scarce drink scarce
       'have nothing to eat or drink'

Coordinating compounds are found throughout the world, with varying semantic relationships between the whole and the parts (Obermüller, 2015). Elaborate expressions are most common in mainland Southeast Asia, where they occupy a position of great prominence. They are often associated with elevated styles of discourse, but they occur in all genres and registers.

Earlier investigators have claimed the order of the constituent words in CCs and EEs in some languages is predictable by rule. Many of the proposed ordering hierarchies are based on phonology (Ting, 1975; Dai, 1986; Mortensen, 2006). Building on this earlier work, Mortensen (2006) posited that Lahu EE orders could be predicted based on vowel quality—like Jingpho (Dai, 1986)—and that Hmong EE orders could be predicted based on tone, echoing earlier claims for Chinese and Qe-Nao (Ting, 1975; Pan and Cao, 1972). These tone and vowel scales were, however, not easy to rationalize in phonetic terms and were used by Mortensen to argue for a phonology in which structure reigns supreme and in which phonetic substance plays only an epiphenomenal role.

---
[1] Code and data available at: https://github.com/dmort27/elab-order

These claims have been viewed with skepticism for two reasons: morphosyntax has been widely seen as providing the inputs to phonology, not being driven by phonology (Chomsky, 1965); and phonology, since Jakobson et al. (1951) and Chomsky and Halle (1968) has usually been seen as grounded in phonetic categories. Some investigators have claimed that sound patterns that are not phonetically natural are inherently unlearnable. They can exist only as linguistic fossils deposited by a history of language change. In this paper, we undertake to investigate what kind of data is needed for (computational) learners to acquire these patterns. We report the following findings:

- Even rather simple classifiers like decision trees can learn to predict the order of EEs in isolation in over 96% of cases (Hmong) and 79% of cases (Lahu) using only phonological information.

- The decision trees for Hmong, Lahu, and Chinese mirror the phonological hierarchies proposed for these languages, suggesting that these hierarchies are empirically robust and learnable from the available evidence.

- However, correct and incorrect orderings of Hmong EEs can be effectively distinguished *in context* by a neural sequence labeling model *without* any phonological information, suggesting that learners would not have to acquire the phonological generalization directly in order to produce well-formed EEs.

## 2 Theoretical Significance

The experiments reported in this paper have a bearing on two assumptions widely held in phonological theory:

1. True phonological generalizations are always grounded in phonetic realities (phonology is natural)
2. Phonology operates on the outputs of syntax and morphology (grammar is serial)

Both of these assumptions have been contested. If the analysis of EE and CC ordering in Mortensen (2006) is sound, neither of these assumptions can be entirely correct.

### 2.1 Phonological patterns and phonetic substance

Starting even before Prague School phonology, it was widely assumed that the grammatical categories and patterns making reference to sound are coherent in terms of physical (articulatory, acoustic, and psychophysical) dimensions. The most common sound patterns found in the world's languages can usually be explained in these terms. For example, in many languages including English, if a nasal is followed by a stop, the place of articulation of the nasal assimilates to that of the stop (*i*[m]*possible* vs. *i*[n]*tolerant* vs. *i*[ŋ]*glorious*).

For phonologically distinctive features, such as those related to place of articulation, this was codified by Jakobson et al. (1951) and injected into generative phonology by Chomsky and Halle (1968). Even more radical statements about the relationship between phonological form and substance have been made since then (Donegan and Stampe, 1979; Flemming, 2013; Hayes, 2011; Donegan and Stampe, 2009; Steriade et al., 2001). While there has never been a complete consensus on the matter (Fudge, 1967; Hyman, 1970; Hale and Reiss, 2000, 2008), it has been widely assumed that phonological patterns that are phonetically incoherent cannot be learned by humans or can be learned only with difficulty (Hayes and White, 2013). For example, Becker et al. (2011) claim that language users do not acquire unnatural statistical patterns that would allow them to distinguish nouns with and without laryngeal alternations between vowel-initial suffixes (while acquiring natural ones). In contrast, Hayes et al. (2009) argue that speakers of Hungarian make use of unnatural patterns in deciding vowel harmony patterns (whether a form ends in a bilabial stop) but have a learning bias towards natural patterns. Artificial grammar learning experiments have been inconclusive but have suggested that the difficult-to-learn phonological patterns are **structurally complex**, not **phonetically unnatural** (Moreton and Pater, 2012a,b).

The phonological ordering generalizations proposed by Mortensen (2006) are structurally quite simple, but often phonetically incoherent. For Hmong EEs, ordering follows the hierarchy presented in Table 1; an EE with an $AB_1AB_2$ form[2] is ordered such that, if $B_1$ and $B_2$ differ in tone, the tone of $B_1$ is higher on the hierarchy than the tone of $B_2$.

---

[2] $AB_1AB_2$ (as in Lahu *chɔ phô? chɔ di*) is also denoted as $ABAC$ in the literature. We use $AB_1AB_2$ in this paper to indicate that the second and fourth words are closely related as they form a potential coordinate compound.

| Order | Orthography | IPA | Description |
|---|---|---|---|
| 1 | -j | ˥˧ | high falling |
| 2 | -b | ˥ | high |
| 3 | -m | ˩̰ | low creaky |
| 4 | -s | ˩ | low |
| 5 | -v | ˧˥ | rising |
| 6 | -g | ˧˩ | falling breathy |
| 7 | -∅ | ˧ | mid |

Table 1: Phonetic values of the tones of Hmong Daw, organized according the the EE ordering scale proposed by Mortensen (2006)

This hierarchy has one phonetically reasonable aspect—the first two tones start high (though their relative rank seems arbitrary). The rest of the hierarchy is puzzling: it goes from lowest to low to rising to falling to neutral. Mortensen's generalization for Lahu elaborate expressions would be easier to reconcile with phonetic substance (the higher in vowel space a vowel is, the better a candidate it is for the first position) were it not that the **best** first-position vowel is /o/, a **mid**, back, rounded vowel. The ordering generalizations that have been proposed for Chinese are similarly arbitrary-looking—they can be stated in terms of historical tonal categories (like the Middle Chinese tones) but appear incoherent in modern lects, in which the phonetic realizations tones have "wandered" phonetically to a dramatic degree.

If it can be shown that these patterns can be learned from naturalistic data, that they are robust predictors of EE and CC ordering, and that models trained to detect correctly ordered EEs and CCs in running text learn to use this kind of phonological evidence to assign labels, it would be suggestive, though not definitive, evidence against the position that phonological constraints must be grounded in phonetic substance.

## 2.2 Word order conditioned on phonology

In mainstream generative linguistics, grammar has usually been viewed as a feed-forward production system. While the nature of this pipeline has changed over various revisions of the theory, a consistent theme is that phonology operates on the output of syntax (Chomsky, 1965, 1981, 1995) and that, therefore, syntax should not be sensitive to phonology. One common version of this theory is the Y model, of which some of the earliest descriptions are found in Chomsky (1981). In this model, surface structure (*Mary hits John* vs. *John is hit by Mary*) is derived first. Then phonetic form (PF; [mɛɹi hɪts dʒɒn]) and logical form (LF; $hit(Mary, John)$) are derived from surface structure. Because surface structure is fixed before PF, syntax should not be sensitive to phonology.[3] In certain other theories of grammar, different levels of representation are computed in parallel and are mutually constraining. An early example of such a framework is Lexical Functional Grammar (Kaplan and Bresnan, 1982; Bresnan et al., 2015). In this class of frameworks, it is expected that phonology *should* be able to influence word order. The question of whether and how phonology can affect word order is significant for larger theories of grammar.

In fact, there is mounting evidence that word order can be sensitive to phonology. It has long been suggested that dative shift in English is sensitive to phonological weight (Ross, 1967) although this claim has also been long contested (Wasow and Arnold, 2003). Some newer evidence comes from coordinate compound and echo reduplication constructions in Japanese, Korean, and Jingpho (Kwon and Masuda, 2019; Dai, 1986). An even more interesting case comes from Tagalog noun-adjective order, which is sometimes viewed as being free but which is actually sensitive to a set of phonological constraints (Shih and Zuraw, 2017). Even more germane to the current discussion are the findings of Benor and Levy (2006) and Morgan and Levy (2016), who found that phonological factors are significant predictors of the sequence of binomial expressions (like *son and daughter*) in English. The current case would enrich the body of relevant evidence in part because, while these cases are all instances of "soft" statistical tendencies, the Hmong ordering generalization is claimed to be nearly categorical (with a few, principled, exceptions).

## 3 Hypotheses

Based on the existing volume of work, we propose the following hypotheses:

---

[3]An important caveat is that—in some versions of generative grammar—syntactic structures are pure hierarchy and are not linearized until PF (phonetic form), when abstract lexical and functional categories are "spelled-out" (Fox and Pesetsky, 2005). This potentially opens the door for interaction between phonology and word order.

1. The order of Hmong and Lahu EEs and Chinese CCs can be predicted phonologically (out of context).

2. The "phonetically unnatural" phonological scales proposed by Mortensen (2006) and Ting (1975) predict the ordering of EEs in Hmong and Lahu and CCs in Chinese (out of context).

3. These scales can be learned by decision tree classifiers (out of context).

4. Phonological information facilitates the recognition of correctly and incorrectly ordered Hmong EEs in context.

## 4 Data

We examine the ordering effects across three languages: Hmong, Lahu, and Chinese (with Middle Chinese and Mandarin pronunciations).

For Hmong, we use a list of 3253 unique elaborate expressions extracted from the 12 million-word Hmong SCH corpus (Mortensen et al., 2022), which was manually annotated and validated by a human expert. All of the EEs are of the form $AB_1AB_2$ where $B_1B_2$ forms a coordinate compound. We also use the entire corpus for the EE tagging task described in Section 5.2. For Lahu, we use a list of 1400 EEs compiled by Matisoff (1989, 2006), which contains both $AB_1AB_2$ and $B_1AB_2A$ forms. For Chinese, we use a list of 254 antonymic coordinate compounds $B_1B_2$ recorded in the *Modern Chinese Dictionary* (Anonymous, 2016). Middle Chinese pronunciations are retrieved from Wiktionary.[4]

## 5 Experiments

### 5.1 Learning Hmong, Lahu, and Chinese CC and EE Ordering with Classifiers over Phonological Features

We first examine whether the orders in elaborate expressions and coordinate compounds can be learned by a classifier. This experiment accomplishes two goals: 1) to reveal the existence and robustness of the patterns in the phonological ordering, and 2) to gain insight into the feature combinations that are most correlated with the ordering effects.

---
[4] Reconstruction from Li (1952)

**Experiment** We use the EE lists described in Section 4 as phrases with the *attested* ordering, and create an *unattested* list of EEs by switching the order of $B_1$ and $B_2$ (occasionally both orders are attested, in which it is not included in the *unattested* list). We then formulate the task as a binary classification problem to predict whether a given ordering is attested or unattested.

To examine the degree to which the order can be predicted by phonology only, we use one-hot features of the onset, rhyme (vowel) and tone constituents in each syllable as classification features. We found that one-hot phonemic features were sufficiently expressive, and that using articulatory features (Mortensen et al., 2016) did not further improve the performance. In Section 5.3 we analyze the effect of adding word embeddings to the feature set. For all classification experiments, we compute the $\chi^2$ statistic on all input features and select the top $K$ features that most correlate with the class label, where $K$ is determined by a development set.

We report the result on two types of classifiers: a decision tree (DT) classifier for maximal interpretability, and a support vector machine (SVM) with RBF kernel for the best classification performance.[5] We also experimented with multi-layer perceptron classifiers of varying widths and depths, but they did not outperform SVM on this dataset. Since other classifiers do not offer the explainability of DT or the performance of SVM, we only report results on these two models.

We split the attested word list into 70%/30% train/test sets before augmenting it with unattested data in order to prevent the same EE from appearing in both the train and test sets. However, it would still be possible for the same $(B_1, B_2)$ to appear in both train and test sets with different $A$ words (repeated words). To eliminate this possibility, we also report results on randomly sampled subsets of EEs wherein all $(B_1, B_2)$ pairs are unique (so that there is no contamination across the train and test sets).

**Rule-Based Classification** We also test how well the ordering scales proposed in Mortensen (2006) perform as a rule-based classifier, compared to a DT and SVM trained on the dataset. This is equivalent to directly examining the distributional patterns of the ordering effects. Table 2

---
[5] Classification models are trained using scikit-learn (Pedregosa et al., 2011)

shows the orders in Hmong, Lahu and Middle Chinese used in the rule-based classifier. When there is a tie, the order is determined randomly.

| Language | Order |
| --- | --- |
| Hmong Tones | j ≺ b ≺ m ≺ s ≺ v ≺ g ≺ ∅ |
| Lahu Rhymes | o ≺ u ≺ i ≺ ɨ ≺ ə ≺ ɔ ≺ e ≺ ɛ ≺ a |
| MC Tones | ping ≺ shang ≺ qu ≺ ru |

Table 2: Linear ordering of tones for rule-based classification, based on Mortensen (2006). $a \prec b$ represents that $a$ occurs before $b$

**Results** Table 3 shows the classification accuracies for all languages. We report results on two classifiers, using two different sets of features: *focal constituent* for the group of phonemes corresponding to the ordering rules (rhyme for Lahu and tone for Hmong and Chinese), and *all constituents* for all the onset, rhyme, and tone phonemes.

We observe a robust correlation between phonology and attested orders in all four languages, as seen by the high accuracy a classifier can attain. Even on unique $(B_1, B_2)$ pairs, the best classifier and feature set achieves 71%–88% accuracy. This means that the ordering effects are not simply due to frequent $(B_1, B_2)$ pairs skewing the statistics; rather, the ordering effect is robust across many $(B_1, B_2)$ pairs in the four languages.

With only the focal constituent feature set, we observe comparable accuracy between the rule-based classification and either statistical classifier. This suggests that the degree to which the focal constituent alone determines EE ordering is no more than the linear ordering scale proposed by Mortensen (2006).[6] However, when phonemes from other constituents are included in the feature set and an SVM is used, we observe an increase of 3–11% in accuracy. This suggests the existence of more complex phonological interactions beyond the linear scale over the focal constituent.

**Visualization of Learned Decision Tree** By examining the learned decision tree, one can derive a linear hierarchy based on the order of features on the *no* branch, and whether each branching action leads to majority attested words or majority unattested words. We find that phonemes that appear topmost in the tree (the most order-defining phonemes) are exactly those at the two ends of the scales proposed by Mortensen (2006), and a decision tree classifier can learn a strikingly similar hierarchy, as shown in Table 4. Details on the derivation and the learned tree are shown in Appendix A.

---
[6] We ran an exhaustive search on all permutations of the tones/vowels, and found the one presented here performs the best as a rule-based classifier.

### 5.2 Learning Hmong EE Ordering as Sequence Labeling

**Experiment** Now we investigate whether models can learn to recognize elaborate expressions and their ordering effects *in context* in a naturalistic corpus. We limit our experiments to Hmong in this section due to the unavailability of EE-annotated corpora in other languages. The Hmong dataset is annotated with `BIO` tags, where a `BIII` sequence represents a labeled EE. We train a neural sequence labeling model to predict the `BIO` tag of each word in a sentence.

We experiment with two types of feature extractors: a bidirectional LSTM and a CNN. We use both word-level and phoneme-level embeddings, following the intuition that the phonologically conditioned ordering helps speakers recognize an EE structure in context. Implementation details and hyperparameters are described in Appendix B.

In addition to the vanilla tagging task, to investigate whether the models can learn the ordering of EEs in context, we perform an experiment where the orders of $B_1$ and $B_2$ are swapped for half of the EEs, and the tags for the swapped EEs are changed to `B-fake` and `I-fake`. This renders the task more difficult as the model needs to both identify an EE in context and classify whether the order has been changed.

To prevent the model from memorizing certain EEs, we split the data into train/val/test sets by partitioning the list of EEs into disjoint sets, so that EEs in the validation and test sets do not appear in the training split. This way, the model is only given unseen EEs at test time. Furthermore, we partition the EEs into swap/no-swap so that occurrences of each EE are either all swapped or all kept unchanged.

**Baseline** The simplest baseline model would be to tag every occurrence of $AB_1AB_2$ (a 4-gram where the first and third words are identical) in the corpus as an EE without any consideration of the word or its phonology. Doing so yields 100% recall but very poor precision, since most occurrences of $AB_1AB_2$ are not elaborate expressions. Three strategies are employed to improve

| Language | Hmong | | Lahu | | Mandarin | Middle Chinese |
|---|---|---|---|---|---|---|
| Data | All | Unique | All | Unique | Unique | Unique |
| N | 6420 | 1404 | 2748 | 1664 | 254 | 251 |
| Rules | 88.8% | 85.5% | 68.3% | 66.3% | – | 70.7% |
| DT (focal constituent) | 89.0% | 85.0% | 67.2% | 64.3% | 65.3% | 70.4% |
| DT (all constituents) | 96.4% | 85.3% | 79.7% | 67.8% | 68.1% | 75.3% |
| SVM (focal constituent) | 89.1% | 85.4% | 67.3% | 64.4% | 65.3% | 70.4% |
| SVM (all constituents) | **96.7%** | **88.3%** | **81.9%** | **71.3%** | **76.1%** | **81.0%** |

Table 3: Classification accuracies with phoneme features (chance is 50%). *Focal constituent* is tone for Hmong and Chinese and rhyme for Lahu. *All constituents* include onset, rhyme and tone. *Unique* for Hmong and Lahu is the average result of 10 randomly sampled subsets of EEs with unique $(B_1, B_2)$. Chinese CCs are always unique.

| Language | | Order |
|---|---|---|
| Hmong | Ling. | j ≺ b ≺ m ≺ s ≺ v ≺ g ≺ ∅ |
| | Tree | j ≺ b ≺ m ≺ v ≺ s ≺ g ≺ ∅ |
| Lahu | Ling. | o ≺ u ≺ i ≺ ɨ ≺ ə ≺ ɔ ≺ e ≺ ɛ ≺ a |
| | Tree | o ≺ u ≺ ... ≺ e ≺ ɔ ≺ ɛ ≺ a |
| MC | Ling. | ping ≺ shang ≺ qu ≺ ru |
| | Tree | ping ≺ shang ≺ qu ≺ ru |

Table 4: Linear orders similar to those posited by Mortensen (2006) are learned by a decision tree

| Model | F1 | Precision | Recall |
|---|---|---|---|
| **Baseline** | 41.32 | 26.15 | 100.00 |
| + regex parsable | 49.24 | 32.83 | 100.00 |
| + wv. sim. thresh | 60.99 | 50.29 | 77.99 |
| + tonal scale | 66.66 | 59.37 | 76.56 |
| **LSTM** | 74.10 | 66.12 | 84.36 |
| + phonemes | 73.14 | 65.39 | 83.09 |
| **LSTM + swap clf.** | 64.38 | 57.54 | 73.29 |
| + phonemes | 63.97 | 56.93 | 73.17 |
| **CNN** | 90.79 | 87.36 | 94.52 |
| + phonemes | 90.26 | 85.98 | 95.58 |
| **CNN + swap clf.** | 89.01 | 85.73 | 92.62 |
| + phonemes | 89.26 | 86.00 | 92.79 |

Table 5: Precision, recall and F1 scores for sequence tagging on the test set. Results are averaged over 9 runs (3 data splits ×3 initial seeds)

the performance of this baseline: (1) ensure that $(A, B_1, B_2)$ are proper Hmong syllables parsable by a regular expression classifier; (2) set a word vector similarity threshold between the two CC words ($B_1$ and $B_2$) so that $cosine(v_{B_1}, v_{B_2}) > \alpha$, since many Hmong EEs have the two CC words of similar meanings (Mortensen, 2006);[7] (3) ensure the tonal scale in Table 2 is followed between $B_1$ and $B_2$

**Results** We report the F1 score of predicted tags on different models in Table 5. For the baseline model, all three strategies improve the tagging performance, suggesting that both semantic similarity and adherence to the tonal scale are indicators of being an EE. Despite the reasonable performance of the baseline, a neural sequence labeling model is able to beat it substantially, achieving a high F1 score in the EE tagging task. In particular, a CNN feature extractor outperforms an LSTM feature extractor. We hypothesize that this is due to a convolution kernel being able to capture non-local interactions in an EE (i.e., identical first and third words, and similar second and fourth words), whereas the linear nature of an LSTM encoder becomes restrictive in this task.

---

[7]Word vectors are trained using Word2Vec (Mikolov et al., 2013). We find that SkipGram outperforms CBOW on this task, hence all results reported are SkipGram embeddings. $\alpha$ is determined by grid search and we find that $\alpha = 0.4$ works best.

When half of the EEs in the form of $AB_1AB_2$ are changed to $AB_2AB_1$ and their tags are modified to `B-fake` and `I-fake` (swap clf. rows in the table), the model is still able to achieve high F1 scores that are only slightly lower than the unswapped counterpart, even though the B and I tags have split into two types. The fact that increasing the number of classes does not degrade the performance very much suggests that the model can learn to distinguish attested and unattested orderings very well. To quantify the model's ability to learn Hmong EE ordering, we calculate an in-context classification accuracy by examining how many correctly identi-fied EEs also have a correct prediction in whether the order has been swapped. We find that the in-context classification accuracy is 99.1% for LSTM and 99.5% for CNN, which are both exceptionally high. Note that this analysis excludes EEs that are not correct identified (both false positives and false negatives). Full confusion matrices are shown in Appendix C.

Interestingly, we find that adding phoneme level features to the input of either LSTM or CNN does not improve the performance in both the swapped and unswapped cases.[8] This result is in contrast with other similar sequence tagging tasks (e.g., NER), where character level features are found to improve performance (Yang et al., 2018; Kuru et al., 2016). More importantly, this result presents a contrast to the robust phonological patterns found in the previous section, as it demonstrates that the model is able to tag elaborate expressions and classify their orders successfully without any reference to phonology. This suggests that the ordering ($B_1$, $B_2$) can be predicted not only via phonology, but also via word-level features through the embeddings trained with the tagging model.

**Visualization of Word Embeddings** It is a rather perplexing result that a tagging model can learn the ordering of EEs via word embeddings only. Figure 1 shows the UMAP projection (McInnes et al., 2018) of two types of learned embeddings into 2D space. Embeddings from the tagging model show clear separation between words that tend to occur first in an EE (in the $B_1$ position) and words that tend to occur second, where as embeddings trained separately on the SkipGram algorithm (Mikolov et al., 2013) show no separation. This suggests that the learned separation is unique to the tagging model. However, there is no way for the model to memorize EEs from the training set, since the test set contains non-overlapping EEs. How, then, would the tagging model learn what words tend to occur first and what words tend to occur second in an EE?

It appears that the model is able to learn the order of $B_1$ and $B_2$ from the occurrences of these *component* words in the training set. For an EE $AB_1AB_2$ in the test set, although $AB_1AB_2$ itself never appears in the training set, $B_1$ and $B_2$ do appear either as a coordinate compound $B_1B_2/B_2B_1$, or as parts of another EE $XB_1XB_2/XB_2XB_1$ in the training set. As shown in Figure 2, appearances of them in the same order greatly outnumbers those of the reversed order. As a result, the model may be able to learn which words tend to be $B_1$ or $B_2$ from these distributional properties of the EE words.

---
[8] We also tried using character features or using only tones (the focal constituent for Hmong), but they were equally ineffective.

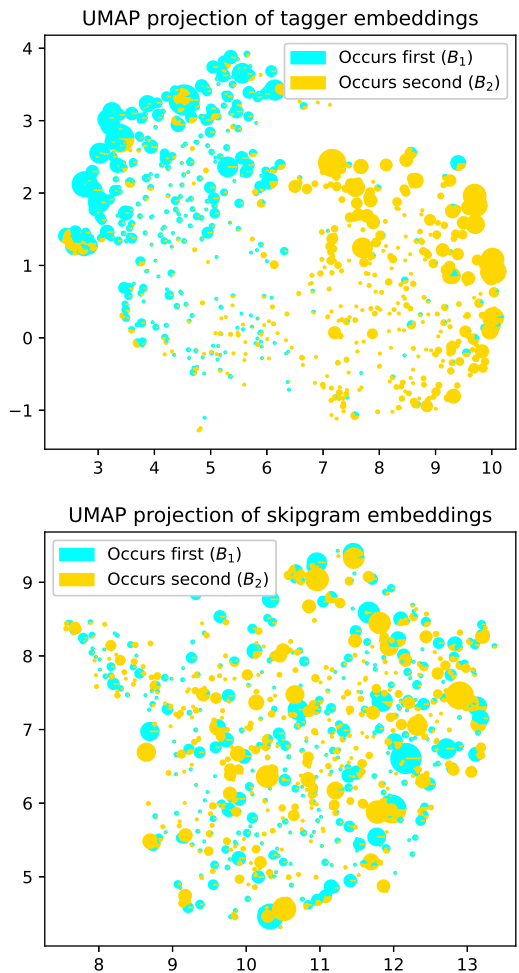

Figure 1: UMAP projection of word embeddings from neural sequence tagger (top) and separately trained SkipGram (bottom). Each circle is one Hmong word. The pie indicates proportion of times the word occurs first or second. Size of the dot indicates frequency in list of EEs.

To further isolate this effect, we perform an experiment where the train/test splits are made so that even component words do no overlap between them (so that the box plot in Figure 2 would be completely empty). We confirm that the tagging performance drop considerably in this setting, with an average F1 score (59.78) unable to beat the strongest baseline. However, even in this setting, we do not find phoneme features to contribute to the tagging performance in a statistically signifi-

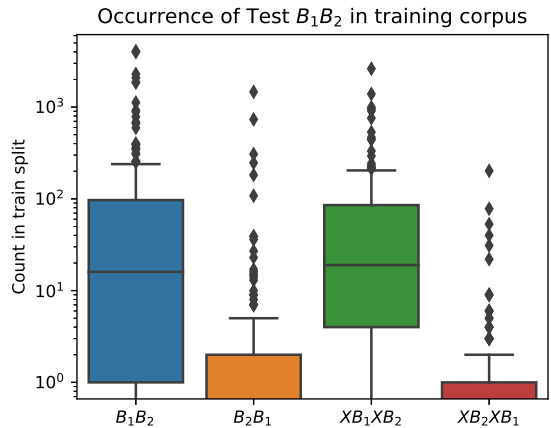

Figure 2: Occurrence of $B_1$ and $B_2$ words in test set EEs ($AB_1 AB_2$) in the training set as different forms ($X \neq A$). Forms with the same ordering (blue and green) outnumbers forms with the reversed ordering (orange and red)

| Data | All | Unique |
| --- | --- | --- |
| focal constituent (tone) | 89.1% | 85.4% |
| all constituents | 96.7% | 88.3% |
| wv-sg | 94.4% | 71.1% |
| wv-tagger | 96.4% | 88.3% |
| all constituents + wv-sg | 96.6% | 88.8% |
| all constituents + wv-tagger | **97.1%** | **93.8%** |

Table 6: Classification accuracies for Hmong using SVM with phoneme and word embedding features. First two rows are from Table 3. *wv-sg*: separately trained skipgram embeddings. *wv-tagger*: embeddings from the CNN sequence tagging model

cant way (F1=61.16, one-sided Wilcoxon signed-rank test p=0.28).

### 5.3 Learning Hmong EE Ordering with Classifiers over Word Vectors

**Experiment** To further investigate to what extent word embeddings determine the order of $(B_1, B_2)$ in Hmong EEs, we revisit the out-of-context classification experiment presented in section 5.1, this time adding word vector features. We experiment with both SkipGram embeddings and embeddings extracted from the CNN sequence tagging model (without swapping). Embeddings from the tagging model are expected to perform better on the classification task, since they are optimized to detect words contained in EEs in the attested order. On the other hand, embeddings separately trained via SkipGram are more "pure," as they only capture the distributional semantics of the words without additional information.

**Results** Table 6 shows the classification accuracies using word embedding features, as well as word embedding combined with one-hot phoneme features. We observe that embeddings trained with the tagger indeed perform better than those trained via SkipGram. What is surprising is that using embedding features from the tagger alone produces a classification accuracy comparable to using all phonemes (88%). Moreover, an even higher accuracy can be achieved by combining phoneme features with embeddings from the tagger. This suggests that EE ordering in Hmong can be predicted from two *independent* but mutually reinforcing routes, namely phonology and lexical distribution. Either method alone is a good predictor of the ordering, but combining the two achieves the best accuracy, because the two routes each offer additional information that are important in predicting the ordering of Hmong EEs.

**Visualization of Feature Importance** With a total 360 features from both phonemes and word vectors[9], we can visualize which features the model find the most important in this classification task by examining the weights learned by the model[10].

Figure 3 shows the proportion of feature types that have the highest importance when varying the number of features ($k$). We see that when $k$ is small, the model overwhelmingly uses phoneme features (especially tones) to perform classification. The test accuracy is impressively 84% with only 12 features – nearly 40% of which are tonal features. As $k$ increases, word embedding features start to gain importance, and the test accuracy can be further improved when word embeddings are incorporated. By the time when $k$ reaches 200, the proportion of each feature type become similar to the natural proportion before selection (dashed lines).

## 6 Discussion

In this paper, we set out to explore the ways that the order of words in EEs and CCs in Hmong, Lahu, and Chinese can be learned by computational models. Motivated by earlier linguists' find-ings, we first use phonological features alone to discriminate between attested and unattested orders of words. We find that in the case of all three languages, the order of words can indeed be predicted phonologically, and the "phonetically unnatural" hierarchies do predict the ordering of EEs and CCs. Furthermore, a decision tree classifier is able to learn more-or-less the same hierarchies, suggesting that speakers of those languages could in principle learn the linear hierarchies through exposure to the language, and use these hierarchies to decide on the correct order of words in EEs and CCs. These findings provide positive evidence for hypotheses 1–3 from Section 3. We then explored the ways models can utilize context and distributional patterns of words to learn the orders in the sequence tagging experiments, and we were not able to find evidence for hypothesis 4. We were surprised to find that models can perform well using only word features, and that adding phonemes to the feature set does not help at all.

---

[9] 58 onsets, 14 rhymes, 8 tones, and 100 word vector dimensions for each of the two words $B_1$ $B_2$.
[10] We switch to an SVM with linear kernel to compare the importance for each feature directly

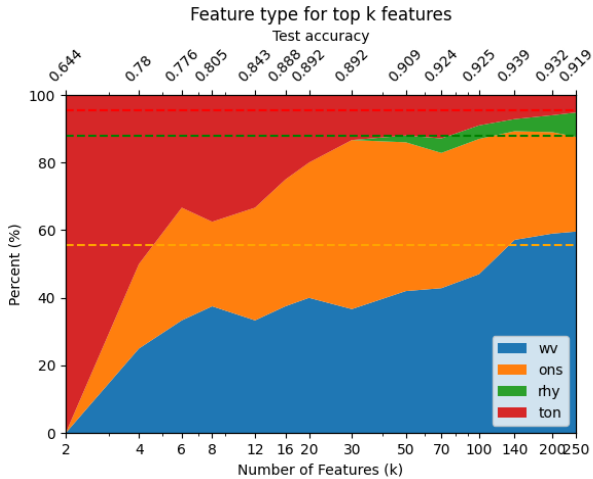

Figure 3: Proportion of feature types for the top k features ranked by importance weight. *x*-axis on the bottom shows the number of features to use on a log scale, while *x*-axis on the top shows the test accuracy for a linear SVM trained on the top *k* features. Dashed lines show the natural proportion of the tone, rhyme, onset and word vector features. Experiment was done with the *Unique* ($B_1$, $B_2$) words.

The seemingly contradictory results of our investigation point in an interesting direction. Information on which a model could rely to learn the ordering of these constructions is present redundantly in phonology (on the one hand) and in lexical and distributional patterns (on the other). When allowed to cooperate on a level playing field, embeddings and phonology-based features both contribute to the identification of well-formed EEs at a similar level. In other words, while it is possible that language users **may** use phonological hierarchies like those proposed in Mortensen (2006) to select appropriate orders for EEs and CCs, it is clearly not the case that they **must** (though they will perform a bit better if they do). These phonological hierarchies may have been more order-defining in the history of the languages, but as the sequence tagging experiments have suggested, they may also have become fossilized in the lexicon and in distributional patterns in the modern form. Many times, a ($B_1$, $B_2$) pair appears abundantly in multiples EEs (as $X B_1 X B_2$), as a CC (as $B_1 B_2$), or in other—more complicated—discourse patterns in the same order, so that language users could learn whether a given word tends to appear in the $B_1$ or $B_2$ position. If a tagging model can learn a word representation that distinguishes between $B_1$ and $B_2$, language users may do the same.

In a sense, these results should be pleasing to both the "structure" (Mortensen, Hale, Reiss) and the "substance" (Hayes, Flemming, Steriade) camps. They show, once again, that generalizations about sounds can be robust but phonetically arbitrary. However, they leave open the possibility that the relevant synchronic generalizations are not actually phonological.

## 7 Future Directions

We have shown two independent routes, namely phonology and lexical distribution, by which computational methods can predict the order of words in Hmong EEs. A language user could probably do the same, relying on both routes to some degree when they need to select the order of words in EEs. However, there is no way to know for sure without conducting a psycholinguistic experiment with native speakers, which would shed light on whether any of the modeling actually translates to human cognition. The Chinese and Lahu cases also raise interesting questions for future work: does the same two-route mechanism work for EEs and CCs in these languages as well? Answering this question will require additional data collection and annotation, but will shed significant light on this theoretically important issue.

## A  Trained Decision Trees

Figures 4, 5, 6 show what the trained decision tree looks like in the three languages. In each tree node, the top half of the box show the current majority class, attested (ATT) or unattested (FAKE), as well as the number of votes. The bottom half of the box shows the variable to branch on. As noted in the main text, a linear ordering can be induced from the tree by following the branches. Take the Hmong tree (Figure 4) as an example. The first factor to split on is *whether $B_1$ has the j tone*, and if the answer is *yes*, the majority of words are attested (255 attested vs 15 unattested/fake). This suggests that *j* has a strong tendency to occur in the $B_1$ position, since it is the most distinguishing factor to split on. Hence *j* can be placed as the first tone on the scale. If $B_1$ does not have the *j* tone, the next question to ask is *whether $B_1$ has the b tone*. Since a *yes* answer again leads to majority attested words (273 attested vs 61 unattested/fake), *b* can be placed second on the scale. The next three questions to ask concern with the $B_2$ word. Since a *yes* answer leads to attested words in all three cases, it suggests that ∅, g and s have a tendency to appear in the $B_2$ position, hence they can be placed on the end of the scale in that order. The next two factors concern with the *j* and *b* tones, which have already been placed on the scale, so we skip them. This process of following the left child (the *no* branch) and placing tones at either end of the scale is repeatedly applied, yielding the induced linear scales shown in Table 4.

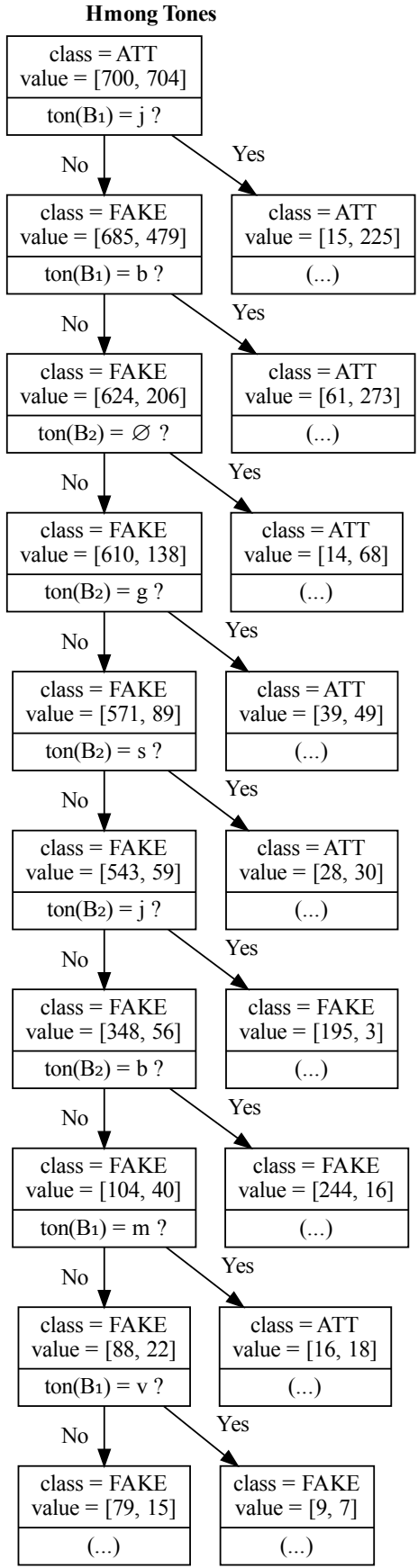

Figure 4: Decision tree trained on Hmong elaborate expressions predicts the following order of tones: j ≺ b ≺ m ≺ v ≺ s ≺ g ≺ ∅ .

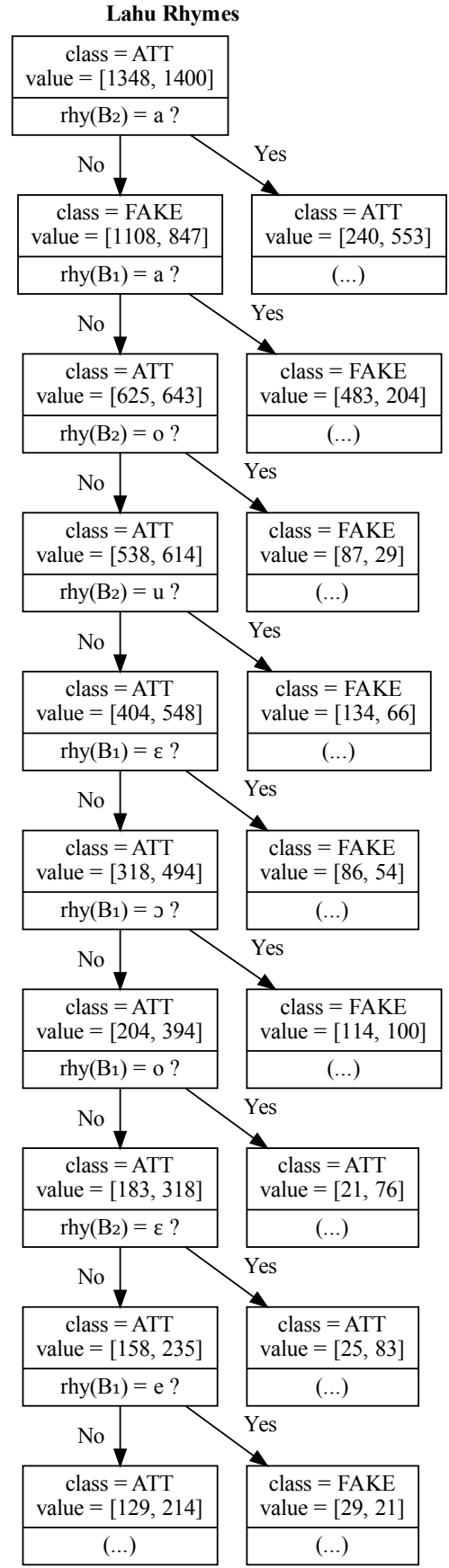

Figure 5: Decision tree trained on Lahu elaborate expressions predicts the following order of rhymes: o ≺ u ≺ ... ≺ e ≺ ɔ ≺ ɛ ≺ a.

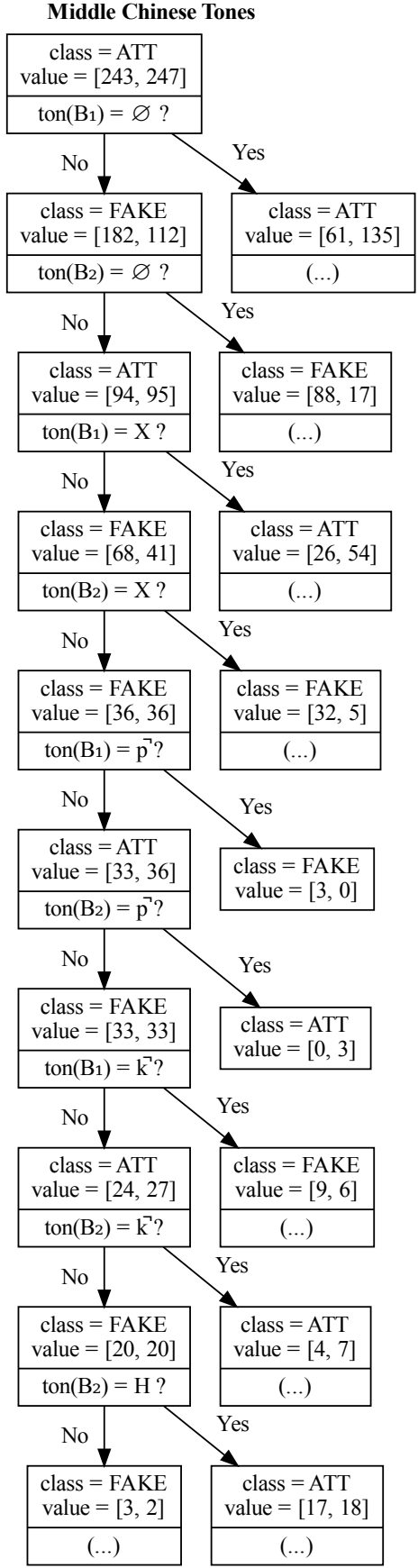

Figure 6: Decision tree trained on Middle Chinese coordinate compounds predicts the following order of tones: ping (∅) ≺ shang (X) ≺ qu (H) ≺ ru ($\vec{p}, \vec{t}, \vec{k}$).

## B Implementation Details

### B.1 Data

The Hmong corpus consists of 740k sentences with a positive rate of around 3.1% (i.e. 96.9% of sentences contain no EEs). The EEs are randomly split into disjoint train and val/test sets with approximate ratios of 91%/4.5%/4.5%. To reduce the possibility that certain splits are easier than others, three such splits are independently produced. The positive sentences are split into train and val/test sets according to the EE partitions, and the negative sentences are split with approximate ratios of 91%/4.5%/4.5%.

### B.2 Models

The sequence tagging model consists of a feature extractor followed by a fully connected layer to predict the tags: {B,I,O} in the unswapped case and {B,B-fake,I,I-fake,O} in the swapped classification experiments. Two feature extractors are used: 1) an LSTM with bidirectional encoding, and 2) a CNN, consisting of four layers of 1D convolution, ReLU, Dropout, and BatchNorm.[11] When character or phoneme level features are used, the character embeddings go through a CharCNN before being concatenated with the word embedding. Details on model configuration is shown in Table 7. The LSTM model contains approximately 1.4M parameters and the CNN contains approximately 1.7M parameters. Our code is based on NCRF++ (Yang and Zhang, 2018).[12]

| Hyperparameter | Value |
| --- | --- |
| Word embed dim | 100 |
| Char embed dim | 30 |
| LSTM hidden dim | 100 |
| CNN hidden dim | 200 |
| CNN kernel size | 3 |
| CharCNN hidden dim | 50 |
| CharCNN kernel size | 3 |
| Dropout probability | 0.5 |

Table 7: Model configuration hyperparameters.

---

[11] An extensive architecture search was not performed, because the purpose of the experiments is not to achieve the best performing model.
[12] https://github.com/jiesutd/NCRFpp, under Apache 2.0 License which permits use for research purposes.

### B.3 Training and Decoding

The model is trained with cross entropy loss using an SGD optimizer with momentum. Early stopping is used on the F1 score of the validation set, with a patience of 10 epochs. During training, negative sentences in the training set are downsampled to 90% (resampled every epoch) instead of 97%, which leads to 3x faster training time but minimal impact on performance. Validation test sets are used in their entirety. Training hyperparameters are shown in Table 8. Training typically takes less than 2 hours to complete on a single GeForce RTX 2080 Ti GPU.

| Hyperparameter | Value |
| --- | --- |
| Batch size | 64 |
| Learning rate | 0.02 |
| SGD momentum | 0.9 |
| Early stopping patience | 10 |

Table 8: Training hyperparameters.

## C Confusion Matrices

Figure 7 shows the confusion matrices for the swap classification experiments. As mentioned in the main text, an in-context classification accuracy can be calculated from the tokens that are correctly identified as part of an EE but may or may not have a correct prediction of the orders (i.e. confused with B-fake). For example, the in-context classification accuracy for the CNN confusion matrix is

$$acc_{CNN} = \frac{439 + 447}{439 + 447 + 4} = 99.55\%$$

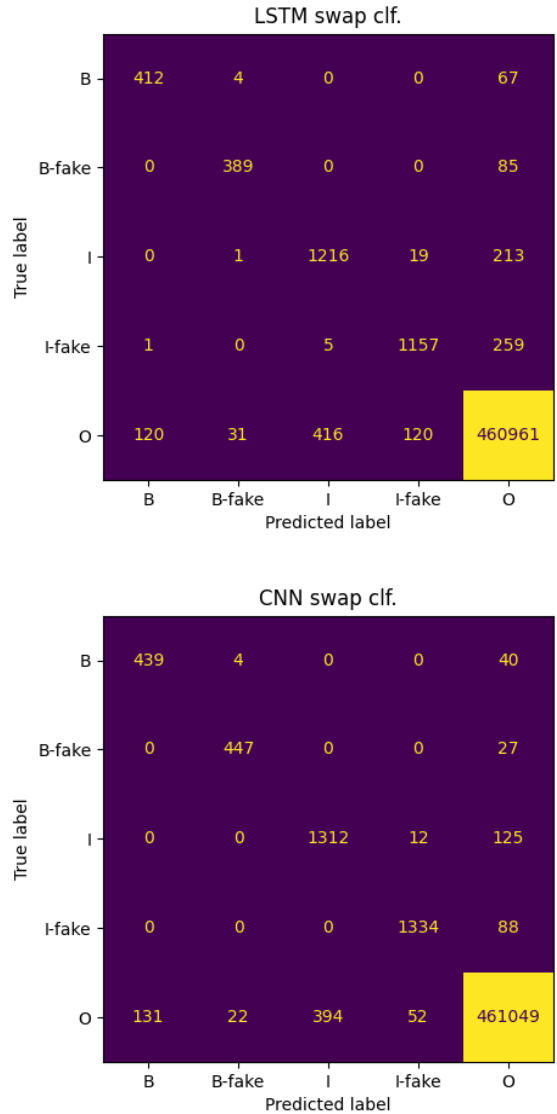

Figure 7: Confusion matrices for the swap classification experiments for LSTM (top) and CNN (bottom)